\pdfoutput=1

\documentclass[11pt]{article}

\usepackage[final]{acl}

\usepackage{times}
\usepackage{latexsym}

\usepackage[T1]{fontenc}

\usepackage[utf8]{inputenc}

\usepackage{microtype}

\usepackage{inconsolata}

\usepackage{graphicx}
\usepackage{cuted}
\usepackage{tabularx}
\usepackage{amssymb}
\usepackage{booktabs}
\usepackage{xcolor} 
\usepackage{adjustbox}
\usepackage{pifont} 
\newcommand{\cmark}{\ding{51}}%
\newcommand{\xmark}{\ding{55}}%
\usepackage{multirow}
\usepackage{amsmath}
\usepackage[utf8]{inputenc}
\usepackage{tcolorbox}
\newcommand{\ourbench}{\textsc{DICE-Bench}}
\newcommand{\ourmetric}{\textsc{DICE-Score}}

\usepackage{colortbl} 
\usepackage{xfp}      
\usepackage{ifthen}   

\title{{\includegraphics[width=0.650cm, height=0.650cm]{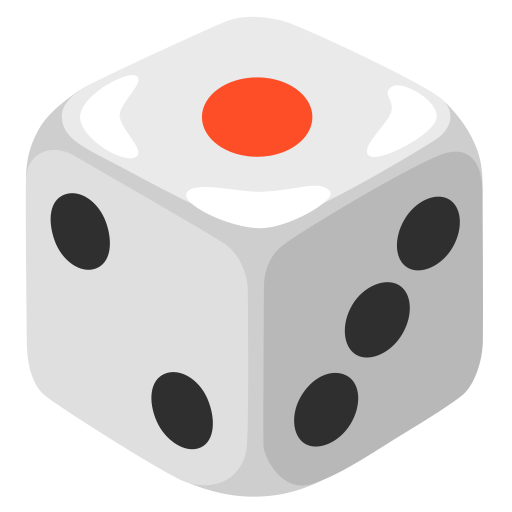}}~\ourbench: Evaluating the Tool-Use Capabilities of\\ Large Language Models in Multi-Round, Multi-Party Dialogues}

\author{
 \textbf{Kyochul Jang\textsuperscript{1}} \quad
 \textbf{Donghyeon Lee\textsuperscript{3, 4}} \quad
 \textbf{Kyusik Kim\textsuperscript{2}} \quad
 \textbf{Dongseok Heo\textsuperscript{1}} \quad\\
 \textbf{Taewhoo Lee\textsuperscript{3, 4}} \quad
 \textbf{Woojeong Kim\textsuperscript{4}} \quad
 \textbf{Bongwon Suh \textsuperscript{1, 2 \thanks{Corresponding author.}}} \quad 
\\[-0.9em]
\\
 \textsuperscript{1}IPAI, Seoul National University \quad \\
 \textsuperscript{2}Department of Intelligence and Information, Seoul National University \quad \\
 \textsuperscript{3}Korea University \quad
 \textsuperscript{4}AIGEN Sciences \quad
 \textsuperscript{5}Cornell University \quad
\\
{\normalsize
\texttt{\{kyochul, kyu823, ty8900, bongwon\}@snu.ac.kr} \quad \texttt{\{dong9733, taewhoo\}}@korea.ac.kr
}
\\
{\normalsize
\texttt{wk247}@cornell.edu}
}

\begin{document}

\maketitle
\renewcommand{\thefootnote}{\textdagger}
\renewcommand{\thefootnote}{\arabic{footnote}} 
\begin{abstract}
Existing function-calling benchmarks focus on single-turn interactions. However, they overlook the complexity of real-world scenarios. To quantify how existing benchmarks address practical applications, we introduce~\ourmetric, a metric that evaluates the dispersion of tool-related information such as function name and parameter values throughout the dialogue. Analyzing existing benchmarks through~\ourmetric~reveals notably low scores, highlighting the need for more realistic scenarios. To address this gap, we present~\ourbench, a framework that constructs practical function-calling datasets by synthesizing conversations through a tool graph that maintains dependencies across rounds and a multi-agent system with distinct personas to enhance dialogue naturalness. The final dataset comprises 1,607 high-\ourmetric~instances. Our experiments on 19 LLMs with~\ourbench~show that significant advances are still required before such models can be deployed effectively in real-world settings. Our code\footnote[1]{\url{https://github.com/snuhcc/DICE-Bench}}, and data\footnote[2]{\url{https://huggingface.co/datasets/OfficerChul/DICE-BENCH}} are all publicly available.

\end{abstract}
\begin{table*}[!t]
\centering
\footnotesize
\begin{tabular}{l c c c c c c}
\toprule
\multirow{2}{*}{\centering \textbf{Benchmark}} 
  & \multirow{2}{*}{\centering \textbf{\# Instances}} 
  & \multicolumn{2}{c}{\textbf{Tool}} 
  & \multicolumn{2}{c}{\textbf{Dialogue}} 
  & \multirow{2}{*}{\centering \textbf{\ourmetric}} \\
\cmidrule(lr){3-4} \cmidrule(lr){5-6}
 & & \textbf{\# Tools} & \textbf{Dependency} 
   & \textbf{Multi-party} & \textbf{Multi-round} &  \\

\midrule
APIBench~\cite{patil_gorilla_2023-1}
  & 17002 & 1645 & \textcolor{red}{\xmark}
  & \textcolor{red}{\xmark} & \textcolor{red}{\xmark} & 0.7895 \\

ToolAlpaca~\cite{tang_toolalpaca_2023}
  & 3938 & 400 & \textcolor{red}{\xmark}
  & \textcolor{red}{\xmark} & \textcolor{red}{\xmark} & 0.5660 \\

ToolLLM~\cite{qin_toolllm_2023-2}
  & 12657 & 16464 & \textcolor{green}{\cmark}
  & \textcolor{red}{\xmark} & \textcolor{red}{\xmark} &  0.5989 \\

ToolBench~\cite{xu_tool_2023-1}
  & 2746 & 8 & \textcolor{green}{\cmark}
  & \textcolor{red}{\xmark} & \textcolor{red}{\xmark} &  0.7225\\

API-Bank~\cite{li_api-bank_2023-1}
  & 2202 & 2211 & \textcolor{red}{\xmark}
  & \textcolor{red}{\xmark} & \textcolor{red}{\xmark} &  1.6318\\

MetaTool~\cite{huang_metatool_2024-1}
  & 21127 & 199 & \textcolor{red}{\xmark}
  & \textcolor{red}{\xmark} & \textcolor{red}{\xmark} &  0.5437\\

TaskBench~\cite{shen_taskbench_2024-1}
  & 17331 & 103 & \textcolor{green}{\cmark}
  & \textcolor{red}{\xmark} & \textcolor{red}{\xmark} &  0.6415 \\

RoTBench~\cite{ye_rotbench_2024-2}
  & 945 & 568 & \textcolor{red}{\xmark}
  & \textcolor{red}{\xmark} & \textcolor{red}{\xmark} &  0.5651\\

\midrule
\textbf{\ourbench~(ours)}
  & \textbf{1607} & \textbf{124} & \textcolor{green}{\cmark} 
  & \textcolor{green}{\cmark} & \textcolor{green}{\cmark} & \textbf{3.6444} \\
\bottomrule
\end{tabular}
\caption{\textbf{Baseline Comparison.} We compare various function-calling benchmark datasets with~\ourbench, demonstrating that~\ourbench~is the only benchmark to encompass both multi-party and multi-round dialogues. We also report~\ourmetric~for every dataset, showing that~\ourbench~handles more realistic tasks.}
\label{tab:comparison}
\vspace{-5mm}
\end{table*}

\section{Introduction}
\label{section:introduction}

Function-calling refers to the ability of LLMs to execute predefined external functions (or APIs) through generating structured calls from natural language input~\cite{qin_tool_2024-1, park_generative_2023-3, GongCoSearchAgentLightweightCollaborativeSearchAgentLargeLanguageModels}. While early virtual assistants (VAs) relied on rigid rule-based systems, LLM-integrated VAs now combine reasoning with external data retrieval~\cite{weizenbaum_elizacomputer_1966}. As interactions grow more complex, there is a growing need for VAs to support multi-party and multi-turn dialogues~\cite{guan_intelligent_2023, VuGPTVoiceTaskerAdvancingMultistepMobileTaskEfficiencyDynamicInterfaceExplorationLearning}.

\begin{figure}[!t]
    \vspace{4mm}
    \centering
    \includegraphics[width=\columnwidth]{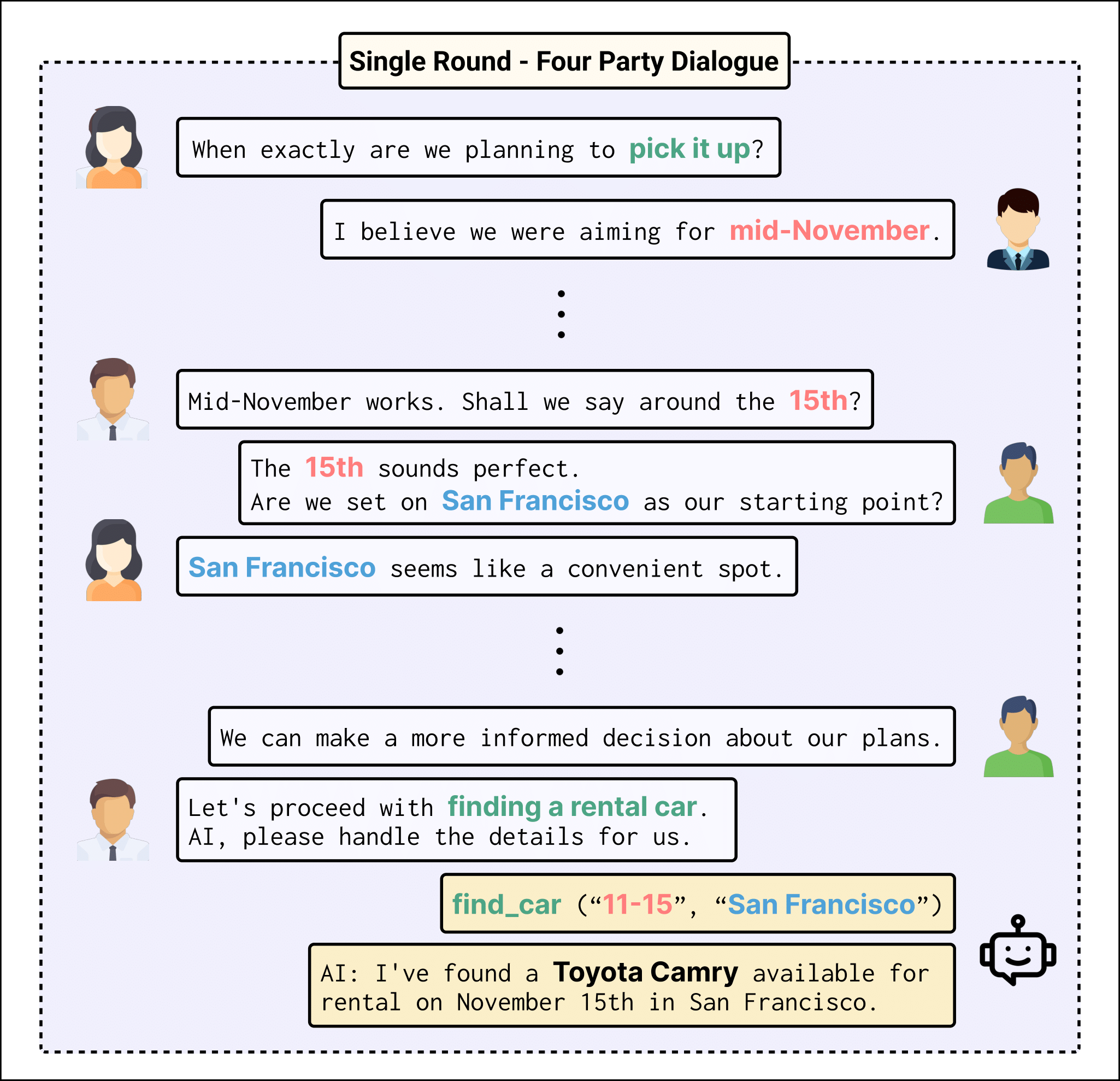}
    \caption{\textbf{Illustration of a Single-Round, Four-Party Dialogue in~\ourbench.} LLMs must identify function-related information from multi-party dialogue. Relevant values in the dialogue are color-coded to match their function call components.}
    \label{fig:singleround}
\end{figure}

Despite advancements, most function-calling benchmarks assume all API parameters are present in a single user utterance, overlooking real-world group chat scenarios~\cite{chen_t-eval_2024-2, zhuang_toolqa_2023-1, basu_api-blend_2024-1}. For example, when people in a group chat decide where to go and which flight to take, a VA must be able to track multiple turns of dialogue to book a hotel and flight ticket. Such complexities remain largely unaddressed by existing benchmarks.

We therefore present \ourbench~(\underline{\textbf{D}}ialogue-based \underline{\textbf{I}}nteractive \underline{\textbf{C}}alling \underline{\textbf{E}}valuation Benchmark), a framework designed to evaluate function-calling performance in realistic multi-party, multi-round dialogues. In our paper, \textit{round} is defined as a complete dialogue cycle consisting of multiple user utterances and system responses, and \textit{dependency} as the condition where the current round's context depends on either the previous round's tool-call output or the content (See Appendix~\ref{fig:multi-round-figure} for illustration of multi-round and dependency). 

In real-world group chats, key details often emerge across multiple turns, requiring accurate tracking for coherent interactions. To address this, we generate diverse dialogues using a multi-agent system, where each agent has a distinct persona. Then, we refine the dataset through automated, rule-based, and human criteria-based filtering. After rigorous validation, our benchmark includes $1,607$ instances covering both single-round and multi-round dialogues. 

Existing benchmarks do not assess function-calling in multi-round, multi-party dialogues, which makes accurate execution challenging due to the tool-related information being dispersed across turns. To quantify this complexity, we propose ~\ourmetric~ (\underline{\textbf{D}}ialogue \underline{\textbf{I}}nformation \underline{\textbf{C}}overage \underline{\textbf{E}}valuation Score), which measures how fragmented tool-related details are within the input context. A higher ~\ourmetric~ indicates greater dispersion, requiring LLMs to integrate scattered information across turns. Experiments on various LLMs show a significant performance drop as ~\ourmetric~ increases, underscoring the need for improved dialogue-tracking and context-integration strategies.

Our contributions are as follows.
\begin{itemize}
    \item To the best of our knowledge, ~\ourbench~is the first multi-round, multi-party benchmark for function-calling, grounded in realistic group chat data and validated through both rule-based and human evaluations.
    \item We introduce the~\ourmetric, a novel metric that captures the complexity of multi-party conversation in the real world by assessing the difficulty of retrieving scattered function call information.
    \item We conducted a thorough evaluation on diverse closed-source and open-source LLMs, analyzing their performance and error cases to provide valuable insights into their limitations in handling fragmented multi-round dialogue contexts.
\end{itemize}

\section{Related Work}
\label{section:related_work}
\begin{figure*}[!t]
    \centering
    \includegraphics[width=\textwidth]{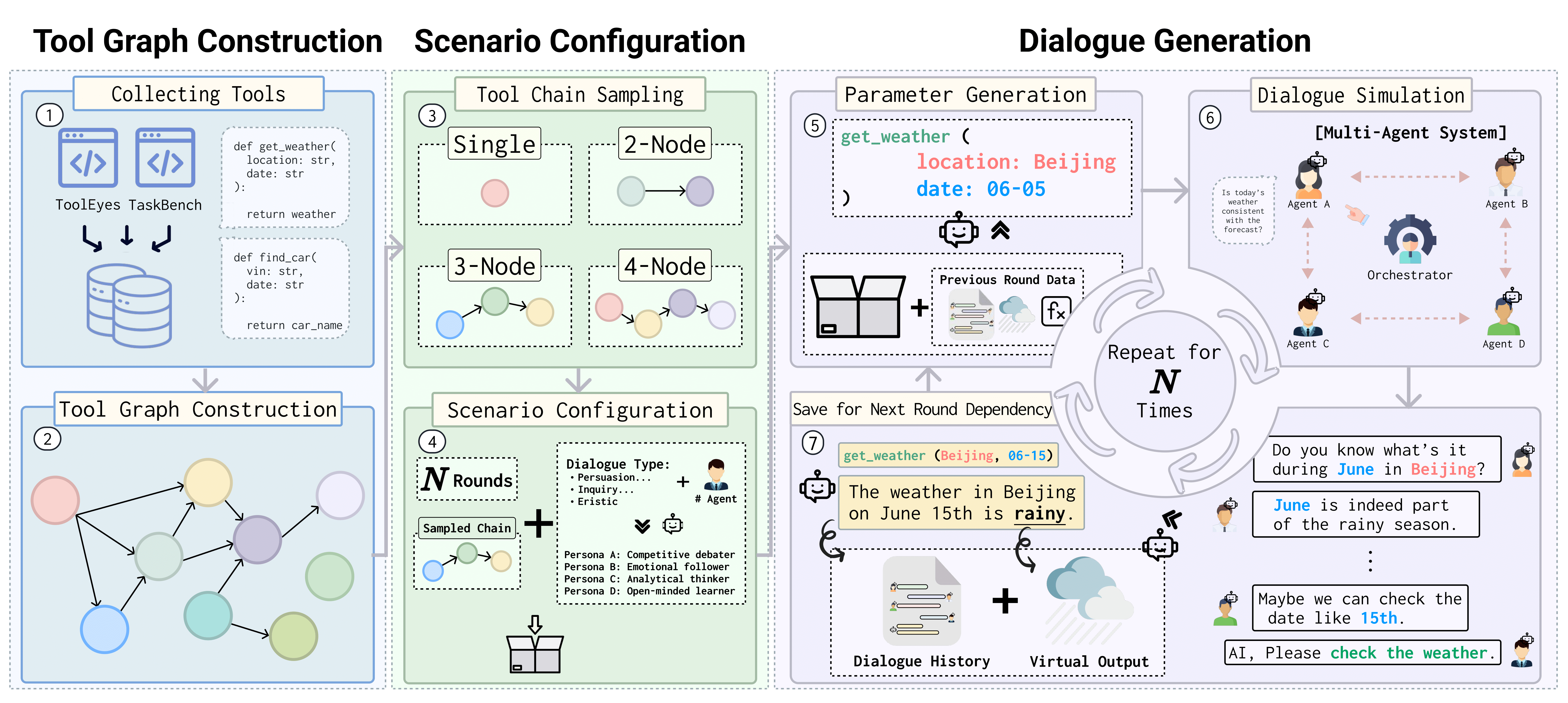}
    \caption{\textbf{\ourbench~data-generation pipeline.} (1) In the \textit{Tool Graph Construction} phase, we build a tool graph from tool collections. (2) In the \textit{Scenario Configuration} step, we sample tool chains and configure dialogue types, personas, and the target number of rounds. (3) In the \textit{Dialogue Simulation} phase, we iteratively generate parameter values for each tool and simulate corresponding multi-party dialogues across N rounds.}
    \label{fig:pipeline}
\end{figure*}

\subsection{Function-Calling Benchmark}
\label{subsection:tool_calling_benchmark}
Recent benchmarks have been developed to evaluate function-calling performance in LLMs~\cite{wang_mtu-bench_2024, kim_seal_2024-1}. Most focus on single-command scenarios~\cite{patil_gorilla_2023-1, huang_metatool_2024-1, qu_tool_2024-1}, while some extend to multi-turn interactions with a single user, increasing task complexity~\cite{li_api-bank_2023-1, wang_mtu-bench_2024, tang_toolalpaca_2023}. However, these approaches overlook the challenges of multi-party dialogues, where tool-related information is distributed across multiple speakers.

Moreover, many existing benchmarks lack rigorous human validation of both tools and instances, leading to datasets that may not reflect real-world conditions~\cite{erdogan_tinyagent_2024, qin_toolllm_2023-2, shen_taskbench_2024-1}. To address these gaps, we introduce ~\ourbench, a benchmark that captures multi-turn, multi-party interactions with comprehensive human validation. Additionally, we propose ~\ourmetric, a metric designed to quantify the dispersion of tool-related information across dialogue contexts, ensuring alignment with real-world complexities.

\subsection{Interactive System and Dialogue}
\label{subsection:interactive_system}
The integration of LLMs into VAs has enhanced their ability to process complex tasks through natural language understanding and reasoning~\cite{sezgin_redefining_2024}. Function-calling further improves this capability by enabling VAs to infer intent before execution, unlike rule-based systems that follow direct commands~\cite{zhang_survey_2025, guan_intelligent_2023, campagna_genie_2019}. As user interactions grow more complex, studies emphasize the need for VAs to handle multi-turn and multi-party dialogues~\cite{abdelaziz_granite-function_2024-1, schick_toolformer_2023, khurana_why_2024}.

Multi-party conversations introduce additional challenges, as they involve diverse dialogue structures shaped by participants’ goals and strategies~\cite{richards_bridging_2025, yeomans_conversational_2022, biber_should_2011, reece_candor_2023}. Academic research categorizes conversations into six types, Persuasion, Inquiry, Discovery, Negotiation, Information-Seeking, Deliberation, and Eristic, each affecting communication complexity differently~\cite{walton_types_2010, walton_commitment_1995}. While function-calling has advanced Human-VA interaction, current benchmarks do not adequately assess multi-party, context-rich dialogues~\cite{inoue_llm_2025, farn_tooltalk_2023}. To address this, we introduce ~\ourbench, a benchmark designed to evaluate LLMs in real-world multi-party interactions.

\section{\ourbench}
\label{section:methods}

In this section, we introduce~\ourbench, a benchmark designed to evaluate the function-calling capabilities of LLMs in multi-round, multi-party dialogues. Unlike previous approaches that concentrate on one-on-one Human-LLM interactions,~\ourbench~ presents dialogue-based inputs in which multiple speakers provide scattered pieces of information over several turns. As shown in Figure~\ref{fig:pipeline}, we also explicitly model inter-round dependencies using Tool Graph. This approach builds upon the concept introduced in TaskBench~\cite{shen_taskbench_2024-1}.

\subsection{Data Construction}
\label{subsection:data_construction}

The data construction phase consists of three main steps: Tool Graph Construction, Scenario Configuration, and Dialogue Generation. Each step undergoes human review and follows clearly defined criteria to ensure the dialogue data is both realistic and consistent.

\paragraph{Tool Graph Construction.}
Our objective is to build dialogue data that mirrors realistic, everyday scenarios where function-calling is needed, such as checking the weather, booking a restaurant, or scheduling events. To achieve this goal, we use the set of tools proposed in the TaskBench~\cite{shen_taskbench_2024-1} and ToolEyes~\cite{ye_tooleyes_2024-1}. We then validate these tools through a combination of manual checks by the authors and LLM-based validation. The two key criteria we used for the filtering are as follows: whether the function calls and parameters realistically reflect daily-life use cases, and whether the collected tools accurately match the intended functions and parameters. After filtering, we construct a Tool Graph to guarantee dependencies between tools.

Formally, we represent our Tool Graph $\mathcal{G}$ as a directed graph $\mathcal{G} = (\mathcal{V}, \mathcal{E})$ where each node $v \in \mathcal{V}$ corresponds to a tool function. A directed edge $(v_i, v_j) \in \mathcal{E}$ signifies that tool $v_j$ depends on the tool $v_i$, either because $v_i$ contains required output or parameters for $v_j$, or because the information produced by $v_i$ is contextually dependent on the execution of $v_j$. Therefore, the structure $\mathcal{G}$ serves as the backbone for multi-round dialogue simulation in a realistic workflow. 

Our Tool Graph consists of 124 nodes and 270 edges, yielding a density of $0.0177$ and an average out-degree of $2.18$. The low density and average out-degree suggest that this graph exhibits a relatively sparse structure, preventing a single function from dominating or becoming overly dependent. This characteristic can offer diverse pathways for automated multi-turn dialogue generation.

\paragraph{Scenario Configuration.}
We integrate various elements to simulate multi-agent, multi-round dialogues in a natural, human-like manner, ensuring each conversation reflects real-world complexity. We begin by sampling tool chains from the Tool Graph, extracting paths ranging from a single node to four nodes, where each node represents a tool per round. For sampling, we employ Depth-First Search (DFS) to enumerate all possible paths, then randomly select the chain. For example, when sampling tools for a two-round dialogue, the sampled tool chain appears as follows: “[$get\_weather$, $book\_hotel$],” meaning the $get\_weather$ function will be used in the first round and the $book\_hotel$ function follows.

Next, we assign a dialogue type based on \citet{walton_commitment_1995}, condensing the seven primary categories into three: persuasion-deliberation-and-negotiation, inquiry-and-information-seeking, and eristic. Although the original reference identifies seven primary types, we merge those that share some similarities. We then vary the number of participants from two to four, spanning a broad complexity range that captures key aspects of real-world multi-party interactions. Lastly, to implement real-world human interactions with distinct personalities, we generate distinct personas for each agent using GPT-4o by leveraging tool information. These configurations cover a broad spectrum of complexity.

\begin{table}[!t]
\centering
\small
\begin{tabular}{c c c c c c}
\toprule
\textbf{Round} & \textbf{Initial} & \textbf{Stage1} & \textbf{Stage2} & \textbf{Stage3} & \textbf{Final} \\
\midrule
1 & 450 & 4 & 7 & 14 & 425 \\
2 & 450 & 5 & 9 & 18 & 418 \\
3 & 450 & 8 & 17 & 26 & 399 \\
4 & 450 & 13 & 11 & 61 & 365 \\
\midrule
\textbf{Total} & 1800 & 30 & 44 & 119 & 1607 \\
\bottomrule
\end{tabular}
\caption{\textbf{Filtering Statistics per Round.} Initial column shows the number of instances before filtering. Stage1–3 show removal counts at each validation step, and Final column shows remaining instances.}
\label{tab:filtering_stats}
\end{table}

\paragraph{Dialogue Generation.}
After preparing essential components, we generate multi-round dialogues in three key steps. First, we perform \textit{Parameter Generation} by prompting an LLM to suggest appropriate parameter values for each tool in the chain. If the current round is not the first round, then we include the conversation history and any previously generated virtual tool-call output to the prompt, ensuring contextual continuity. 

Next, we carry out \textit{Dialogue Simulation} using a multi-agent system. Each agent has a distinct persona, and an orchestrator dynamically regulates turn-taking based on the evolving conversation flow. This setup emulates real-world multi-party conversations. Finally, at the end of each round, we store the dialogue along with any generated virtual outputs, which serve as a context for the next round’s parameter generation. We repeat this process $N$ times, where $N$ is the length of the chain. Using this approach, we produced a total of 1,800 ($450 \times 4$) dialogues across four rounds.

\subsection{Validation Pipeline}
\label{subsec:validation_pipeline}

We employ a three-stage filtering process to convert the raw dialogues into high-quality data. After the first automated stage, each subsequent filtering step involves human validation to ensure that the final dataset meets our criteria for realism, coherence, and functional correctness.

\paragraph{Stage 1: Automatic Evaluation.}
In the initial stage, we use G-Eval~\cite{liu_g-eval_2023-1} with GPT-4o to evaluate each dialogue according to six criteria: Coherence, Consistency, Fluency, Human-likeness, Persona Consistency, and Relevance. Each criterion is rated on a 5-point Likert scale. Although model-based evaluation may introduce certain biases,~\citet{liu_g-eval_2023-1}~have shown a high Spearman correlation between automated scores and human judgments. We then prompt GPT-4o to classify each dialogue into one of the three designated dialogue types. We remove it if a dialogue’s average G-Eval score falls below 4.0 and is assigned an incorrect type.

\paragraph{Stage 2: Rule-Based Filtering.}
Following the automatic evaluation, we discard dialogues that violate explicit rules. First, any conversation containing GPT-generated refusals (e.g., “I’m sorry, but…”) is removed. Second, we check if at least one user turns explicitly or implicitly addresses an “AI” or “Assistant”. In ambiguous cases, authors revisit each dialogue to confirm whether indirect requests, such as rhetorical questions to AI, are being made.

\paragraph{Stage 3: Criteria-Based Filtering.}
In the final stage, all authors evaluate each remaining dialogue across three dimensions: Conversation Quality, Functional Integration, and Real-World Applicability. Detailed guidelines are provided in the Appendix~\ref{sec:appendixValidationCriteria}. These dimensions encompass 15 sub-criteria in total, with seven dedicated to conversation quality, five to function integration, and three to overall realism. We remove the instance if a dialogue scores below 10 out of 15. 

These three filtering stages produce a curated dataset that maintains coherence and accurately represents challenging function call scenarios. In Table~\ref{tab:filtering_stats}, we describe the number of data points that were eliminated at each filtering stage and the number that eventually remained in the final dataset.

\subsection{Task Setup and Benchmark Structure}
\label{subsec:task_setup}
In this section, we explain how our benchmark is structured, and describe our overall task setup. Specifically, we illustrate how multi-round, multi-party dialogues challenge LLMs to aggregate scattered information and perform accurate function calls.

\subsubsection{Benchmark Structure}
\label{subsubsec:benchmark_structure}
Our dataset comprises four rounds, ranging from Round 1 to Round 4. Each round progressively increases in complexity by expanding the contextual scope and requiring the model to handle diverse personas and manage rapid context shifts, from two participants up to four participants. We also include three distinct dialogue styles to mirror varied real-world scenarios.

\begin{table}[!t]
\centering
\setlength{\tabcolsep}{4pt}
\begin{tabular}{c|cccc}
\hline
\textbf{Index} & \textbf{1} & \textbf{2} & \textbf{3} & \textbf{4} \\
\hline
Round          & 425        & 418        & 399        & 365        \\
Party          & -          & 569        & 519        & 519        \\
Dialogue Type             & 545        & 549        & 513        & -          \\

\hline
\end{tabular}
\caption{\textbf{Data Statistics of~\ourbench.} For the Dialogue Type row, indices 1-3 correspond to “Eristic”, “Persuasion, Deliberation and Negotiation”, and “Inquiry and Information Seeking”, respectively.}
\label{tab:data_stats}
\end{table}

We generate 50 dialogues per round for each of the three-party configurations and three dialogue types, yielding 450 dialogues$(450 = 50 * 3 * 3)$ per round. With 4 rounds, this results in a total of $1,800$ dialogues$(1800 = 450 * 4)$ overall. After 193 dialogues are removed through the validation pipeline, we obtain $1,607$ final instances. Refer to Table~\ref{tab:data_stats} for detailed data statistics for each configuration.

\subsubsection{Task Setup}
\label{subsubsec:task_setup}
In DICE-Bench, our aim is to evaluate how well LLMs can perform function-calling under realistic multi-party dialogue conditions. Therefore, we need to inference LLMs on our synthesized dialogue datasets. The input consists of a multi-round multi-party dialogue, and collected tool documents from \textit{Tool Graph Construction} phase. The three types of input are fed to the target LLMs as a hard prompt. We define the task as identifying the exact function name and parameter values based on the given user instruction and dialogue. Thus, the benchmark tests the model's ability to (i) identify the appropriate function among available tools, and (ii) extract or synthesize the correct parameter values within the given conversation. This setup more closely aligns with real-world Human-VA interactions, where relevant context is often distributed throughout extended dialogues rather than being neatly encapsulated in a single instruction.

\begin{figure}[!t]
    \centering
    \includegraphics[width=\columnwidth]{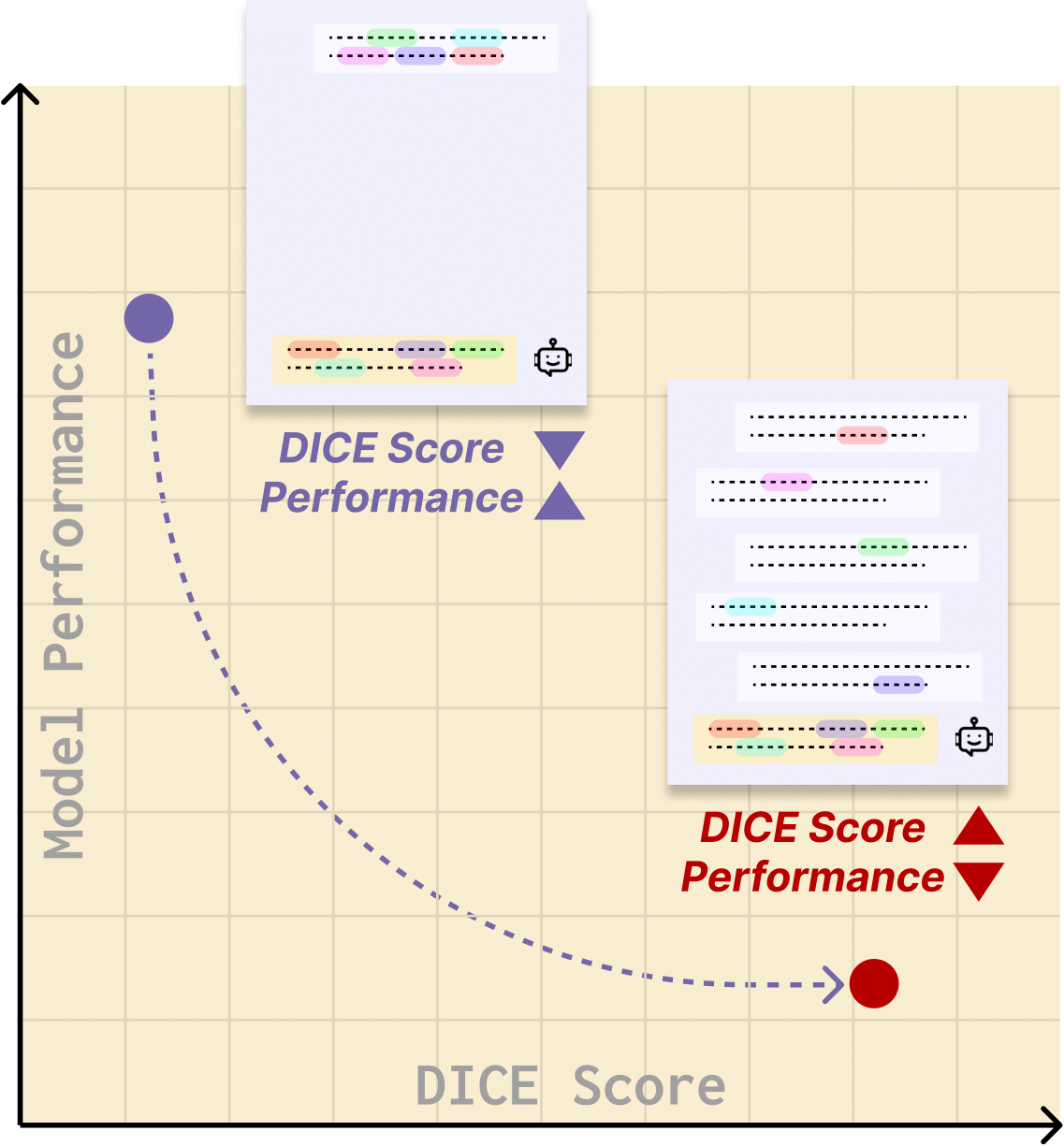}
    \caption{\textbf{Inverse Correlation between~\ourmetric~and Model Performance.} Lower~\ourmetric~indicates that the input instruction is more challenging, suggesting that the LLM is capable of handling complex scenarios.}
    \label{fig:dice_inverse_correlation}
\end{figure}

\subsection{~\ourmetric}
\label{subsection:DICE_metric}

We propose~\ourmetric~to quantify how difficult the given input is for function-calling across existing benchmark datasets as they do not fully reflect practical situations. However, the lack of a metric to measure this aspect is hindering the progress towards more challenging tasks. Although some studies have discussed the notion of information coverage by quantifying how much of the  input context is necessary for answering queries, none have proposed a metric that explicitly captures how dispersed or fragmented these details are within a dialogue for function-calling tasks. Specifically, according to~\citet{goldman_is_2024}, "scope" is defined as "how much required data can be found", but does not formalize a direct metric. Also, the existing long-context coverage method~\citet{lee_ethic_2024} measures how dense the information is distributed throughout the long context, rather than quantifying its sparseness across multiple utterances. 

To address this gap, we introduce ~\ourmetric~(\underline{\textbf{D}}ialogue \underline{\textbf{I}}nformation \underline{\textbf{C}}overage \underline{\textbf{E}}valuation Score), a metric that assesses how challenging it is to perform a function call within a given context by estimating the distribution of tool-related knowledge. We designed~\ourmetric~to yield higher scores when there is a large amount of function-related information to identify, but also when this information is distributed sparsely and non-repetitively. This, in turn, makes it more difficult for LLMs to locate the necessary information. Formally, we define the DICE metric as follows:

\begin{equation}
\begingroup
\text{DICE}(S,T) =
\frac{\min\Big( |S_{\neq 0}|, T \Big) \cdot \sqrt{|S| \cdot T}}
{\sum_{i \in S} \ln(1 + \alpha \times S_i)}.
\label{eq:dice_formula}
\endgroup
\end{equation}

\noindent
\paragraph{Notation.}
Let the dialogue consist of $n$ utterances, and define $S = (S_1, \dots, S_n)$ as a vector where each $S_i$ indicates the number of function-related items mentioned in the $i$-th utterance.  
Removing all zero entries from \(S\) yields the subsequence \(S_{\neq 0}\); therefore \(|S_{\neq 0}|\) equals the number of utterances that mention at least one such item. \(T\) denotes the total number of distinct function-related items that must be identified across the entire dialogue. For example, if the ground truth function-call is $book\_hotel(Vienna, Austria, 07-27)$, then $T=4$, comprising one for the $book\_hotel$ and three for its arguments: $Vienna, Austria$, and $07-27$. $\alpha$ is a positive constant to control a penalty for repeated mentions of the same items. We set \(\alpha = e^2\), which ensures in the boundary case \(T=|S_{\neq0}|=1\) that the~\ourmetric~remains strictly increasing.

\paragraph{Key Properties.}
To obtain \(S_i\) in practice, we employ a custom prompt to GPT-4o-mini (details in Appendix~\ref{prompt:si_prompt}). We highlight four key properties of~\ourmetric:

\begin{enumerate}
    \item \textbf{Coverage vs.\ Dispersal:}\\
    The term \(\min(|S_{\neq 0}|, T)\) rewards spreading items across dialogue turns, aligning with studies on information dispersion in corpus linguistics and multi-turn dialogue systems \cite{manning1999foundations, jurafsky2019speech}.
    
    \item \textbf{Discouraging Redundancy:}\\
    The logarithmic penalty \(\sum_{i \in S} \ln(1+\alpha \times S_i)\) downweights repeated mentions, similar to TF-IDF weighting in information retrieval \cite{salton1988term}.
    
    \item \textbf{Scale Adjustment:}\\
    The factor \(\sqrt{|S| \times T}\) normalizes the score with respect to dialogue length and item count, analogous to cosine normalization in document similarity \cite{manning1999foundations}.
    \item \textbf{Balanced Realism:} Repeating the same items in every utterance increases the denominator, lowering~\ourmetric, while mentioning items too sparsely keeps the numerator small. Thus, a high~\ourmetric~ indicates that items are well-distributed across the conversation. Moreover, when the utterance count \( t \) and item repetition remain fixed (i.e., \( T \) is proportional to \( S_i \) for \( S_i \geq 1 \)), we show (Appendix~\ref{appendix:alpha_proof}) that there exists \( \alpha \) with \( e^2 \leq \alpha \) such that the~\ourmetric~strictly increases with the number of distinct tools.
\end{enumerate}

\begin{figure}[!t]
    \centering
    \includegraphics[width=\columnwidth]{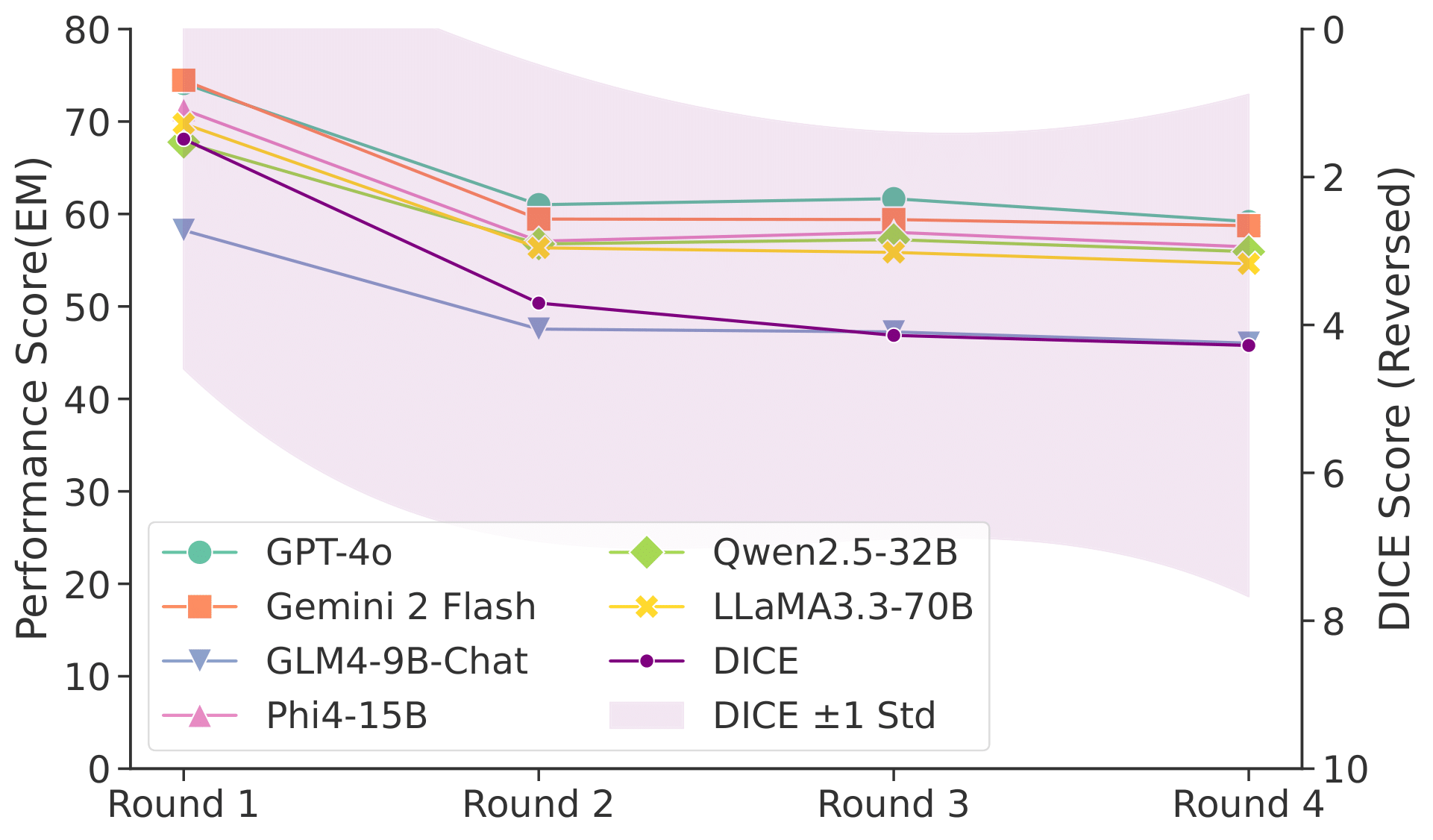}
    \caption{\textbf{EM Performance Scores vs~\ourmetric.}~\ourmetric~has been inverted to highlight its correlation with LLMs performance. The "DICE" in the legend represents the~\ourmetric, and the purple-shaded region indicates \(\pm 1\) standard deviation of ~\ourmetric.}
    \label{fig:dice_alignment}
\end{figure}

\paragraph{Alignment between~\ourmetric~and Human Performance.}

To validate \ourmetric, we measured the Spearman Correlation between the human performance on~\ourbench~and~\ourmetric~using a statistically grounded subset of 311 samples. This subset size was determined based on a 95\% confidence level, a 5\% margin of error, and a conservative estimate of maximum variability ($p = 0.5$). The calculation incorporated a Finite Population Correction (FPC) to account for the dataset's finite size.

Samples were proportionally drawn from four rounds of data, 425, 418, 399, and 365 samples in Rounds 1 to 4, resulting in evaluation subsets of 82, 81, 77, and 71 samples, respectively. Human participants completed function-calling tasks for each round in the sample, achieving accuracies of 80.5\%, 69.1\%, 51.9\%, and 49.3\%. Corresponding values of~\ourmetric, which reflect increasing task difficulty, were 1.42, 3.25, 4.55, and 5.36. This statistics are summarized in Table~\ref{tab:human_alignment}.

To assess the alignment between human performance and the proposed difficulty metric, we computed the Pearson correlation coefficient. The analysis revealed a strong negative correlation ($r \approx -0.984$), indicating that higher~\ourmetric~values were associated with lower human accuracy. This trend is consistent across rounds, from 80.5\% accuracy at $DICE=1.42$ (Round 1) to 49.3\% at $DICE=5.36$ (Round 4). A t-test confirmed the statistical significance of this correlation, yielding a t-value of approximately $-7.81$ ($p < 0.05$, 2 degrees of freedom).

These results demonstrate that~\ourmetric~effectively captures the difficulty of input dataset, with both human evaluation and statistical analysis supporting its validity. Please refer to Appendix~\ref{appendix:human_alignment_calculation} for calculation details. Moreover, in Appendix~\ref{appendix:alpha_proof}, we show how~\ourmetric~performs as expected when tool-related items increase, as long as dispersal and repetition remain balanced. A higher~\ourmetric~means crucial information is spread over multiple turns. Lastly, in Figure~\ref{fig:dice_alignment}, we illustrate how~\ourmetric~correlates with the model performance, and Table~\ref{tab:comparison} compares~\ourmetric~across various function-calling benchmarks.

\begin{table}[t!]
    \centering
    \begin{tabular}{lccc}
        \toprule
        Round & $N$ & EM (\%) & \ourmetric \\
        \midrule
        1 & 82 & 80.5 & 1.42 \\
        2 & 81 & 69.1 & 3.25 \\
        3 & 77 & 51.9 & 4.55 \\
        4 & 71 & 49.3 & 5.36 \\
        \bottomrule
    \end{tabular}
    \caption{Human Evaluation Results by Round. EM (\%) denotes the Human performance  using EM metrics on~\ourbench~ samples. $N$ refers to the sample size per round.~\ourmetric column refers to the~\ourmetric~for each round on the entire dataset.}
    \label{tab:human_alignment}
\end{table}
\newcommand{\colorHighVal}[1]{%
  \begingroup
    \setlength{\fboxsep}{3pt}%
    \edef\fillratio{\fpeval{20}}
    \colorbox{green!\fillratio}{\textcolor{black}{#1}}%
  \endgroup
}
\newcommand{\colorLowestVal}[1]{%
  \begingroup
    \setlength{\fboxsep}{3pt}%
    \edef\fillratio{\fpeval{30}}
    \colorbox{red!\fillratio}{\textcolor{black}{#1}}%
  \endgroup
}
\newcommand{\colorHighestOS}[1]{\colorHighVal{#1}}

\begin{table*}[!t]
\centering
\renewcommand{\arraystretch}{1.15}
\resizebox{\textwidth}{!}{%
\begin{tabular}{c c | c c c c c | c c c c}
\toprule
\multirow{2}{*}{\textbf{Category}} & 
\multirow{2}{*}{\textbf{Model}} &
\multicolumn{5}{c|}{\textbf{Round}} &
\multicolumn{4}{c}{\textbf{Party}} \\
\cmidrule(lr){3-7}\cmidrule(lr){8-11}
 & 
 & \textbf{R1} & \textbf{R2} & \textbf{R3} & \textbf{R4} & \textbf{Avg(R)}
 & \textbf{P2} & \textbf{P3} & \textbf{P4} & \textbf{Avg(P)} \\
\midrule

\multirow{4}{*}{\textbf{Closed-Source}}

& GPT-4o
  & 74.1176
  & \colorHighVal{61.0048}
  & \colorHighVal{61.6541}
  & \colorHighVal{59.1781}
  & \colorHighVal{63.9887}
  & \colorHighVal{61.2045}
  & \colorHighVal{62.2997}
  & \colorHighVal{62.4396}
  & \colorHighVal{61.9813}
  \\

& GPT-4o-mini
  & \colorLowestVal{66.8235}
  & 57.9545
  & 57.8947
  & 56.7123
  & \colorLowestVal{59.8463}
  & \colorLowestVal{57.5280}
  & \colorLowestVal{58.5337}
  & 59.3800
  & 58.4806
  \\

& Gemini 2 Flash
  & \colorHighVal{74.4706}
  & 59.4498
  & 59.3985
  & 58.7329
  & 63.0129
  & 59.6989
  & 61.1779
  & 61.6747
  & 60.8505
  \\

& Gemini 2 Flash Lite
  & 70.9412
  & \colorLowestVal{56.8182}
  & \colorLowestVal{57.3517}
  & \colorLowestVal{56.6781}
  & 60.4473
  & 58.3333
  & 58.5737
  & \colorLowestVal{58.4944}
  & \colorLowestVal{58.4671}
  \\
\midrule

\multirow{8}{*}{\shortstack{\textbf{Open-Source}\\[3pt]\footnotesize{(7B -- 9B)}}}

& Qwen2.5-7B
  & 53.0588
  & 40.1316
  & 37.9282
  & 36.7123
  & 41.9577
  & 39.0056
  & 40.3045
  & 39.5330
  & 39.6144
  \\

& Mistral-7B
  & 50.3529
  & 38.8158
  & 35.2130
  & 33.3219
  & 39.4259
  & 36.7997
  & 37.2997
  & 36.6747
  & 36.9247
  \\

& Hammer-2.1-7B
  & 31.2941
  & 22.1292
  & 19.4653
  & 17.8425
  & 22.6828
  & 20.7633
  & 20.4728
  & 20.8937
  & 20.7099
  \\

& EXAONE-3.5-7.8B
  & \colorLowestVal{1.8824}
  & \colorLowestVal{0.3589}
  & \colorLowestVal{0.2089}
  & \colorLowestVal{0.3767}
  & \colorLowestVal{0.7067}
  & \colorLowestVal{0.4902}
  & \colorLowestVal{0.5609}
  & \colorLowestVal{0.4026}
  & \colorLowestVal{0.4846}
  \\

& LLaMA3.1-8B
  & 26.3529
  & 19.6172
  & 15.3718
  & 15.0685
  & 19.1026
  & 16.4566
  & 17.5080
  & 18.2367
  & 17.4004
  \\

& CALM-8B
  & 2.8235
  & 4.0072
  & 3.5505
  & 2.3973
  & 3.1946
  & 2.8361
  & 3.6058
  & 3.0193
  & 3.1537
  \\

& ToolAce-8B
  & 2.4706
  & 0.6579
  & 0.3342
  & 0.5137
  & 0.9941
  & 0.7003
  & 0.8013
  & 0.6039
  & 0.7018
  \\

& GLM4-9B-Chat
  & \colorHighestOS{58.2353}
  & \colorHighestOS{47.5478}
  & \colorHighestOS{47.2431}
  & \colorHighestOS{46.0274}
  & \colorHighestOS{49.7634}
  & \colorHighestOS{47.6190}
  & \colorHighestOS{47.2756}
  & \colorHighestOS{49.3156}
  & \colorHighestOS{48.0701}
  \\
\midrule

\multirow{4}{*}{\shortstack{\textbf{Open-Source}\\[3pt]\footnotesize{(13B -- 20B)}}}

& NexusRaven-V2-13B
  & \colorLowestVal{34.2353}
  & \colorLowestVal{24.1627}
  & \colorLowestVal{20.7602}
  & 20.7192
  & \colorLowestVal{24.9693}
  & \colorLowestVal{23.0742}
  & \colorLowestVal{22.6763}
  & \colorLowestVal{23.0274}
  & \colorLowestVal{22.9260}
  \\

& Qwen2.5-14B
  & 58.3529
  & 48.8636
  & 49.1646
  & 47.2945
  & 50.9189
  & 50.0700
  & 48.9183
  & 49.1143
  & 49.3675
  \\

& Phi4-15B
  & \colorHighestOS{71.2941}
  & \colorHighestOS{57.0574}
  & \colorHighestOS{58.0201}
  & \colorHighestOS{56.4384}
  & \colorHighestOS{60.7025}
  & \colorHighestOS{57.4580}
  & \colorHighestOS{58.6538}
  & \colorHighestOS{60.0644}
  & \colorHighestOS{58.7254}
  \\

& Granite-20B
  & 58.7059
  & 31.6986
  & 24.8120
  & \colorLowestVal{19.2808}
  & 33.6243
  & 27.8711
  & 28.5657
  & 27.2544
  & 27.8971
  \\
\midrule

\multirow{3}{*}{\shortstack{\textbf{Open-Source}\\[3pt]\footnotesize{(32B -- 70B)}}}

& Qwen2.5-32B
  & 67.7647
  & \colorHighestOS{56.7584}
  & \colorHighestOS{57.2264}
  & \colorHighestOS{55.9247}
  & \colorHighestOS{59.4185}
  & \colorHighestOS{57.5280}
  & \colorHighestOS{57.4920}
  & 58.3736
  & \colorHighestOS{57.7979}
  \\

& LLaMA3.3-70B
  & \colorHighestOS{69.7647}
  & 56.3397
  & 55.8480
  & 54.6233
  & 59.1439
  & 55.9524
  & 56.7708
  & \colorHighestOS{58.4541}
  & 57.0591
  \\

& CALM-70B
  & \colorLowestVal{41.2941}
  & \colorLowestVal{36.3636}
  & \colorLowestVal{40.2256}
  & \colorLowestVal{38.7671}
  & \colorLowestVal{39.1626}
  & \colorLowestVal{38.1653}
  & \colorLowestVal{38.9423}
  & \colorLowestVal{39.9356}
  & \colorLowestVal{39.0144}
  \\

\bottomrule
\end{tabular}%
}
\caption{\textbf{Main Experiment Results of \ourbench.} Reported scores are EM (Exact Match) scores. For each block, the single highest 
(green) and lowest (red) values are highlighted \emph{within that block only}. See Section~\ref{sec:experiments}~for more details.}

\label{tab:main_experiments}
\end{table*}

\section{Experiments}
\label{sec:experiments}

\subsection{Model Selection}
\label{subsec:model_selection}

We evaluated a total of 19 LLMs that support at least 8k context window size. Also we excluded the reasoning models. The closed-source cohort includes GPT-4o and GPT-4o-mini~\cite{openai_gpt-4_2024-2}, along with Gemini 2 Flash and Gemini 2 Flash Lite~\cite{team_gemini_2024}. Meanwhile, our open-source lineup spans a wide range of general-purpose models, including LLaMA3~\cite{touvron_llama_2023}, Qwen2.5~\cite{qwen_qwen25_2025}, Mistral~\cite{jiang_mistral_2023}, 
EXAONE~\cite{research_exaone_2024},
Phi4~\cite{abdin_phi-4_2024},
GLM4-Chat~\cite{glm_chatglm_2024}. In addition, we evaluate tool-specific models that have been fine-tuned on tool datasets, including
Hammer2.1~\cite{wang_hammerbench_2024}, ToolAce~\cite{liu_toolace_2024-1},  CALM~\cite{acikgoz_can_2025}, NexusRaven-V2~\cite{nexusraven}, Granite~\cite{abdelaziz_granite-function_2024-1}.

\subsection{Evaluation Metrics}
\label{subsec:evaluation_metrics}
Since our benchmark aims to evaluate LLM tool-calling performance under multi-round and multi-party input scenarios, we divided the assessment into four-round and three-party configurations. To measure performance, we adopt the Exact Match (EM) metric, which evaluates whether the LLM selects the exact function along with its corresponding parameters. The final score is obtained by averaging the EM across the configuration dataset.

\subsection{Experimental Findings}
\label{subsec:experimental_findings}

\subsubsection{Results}
\label{subsubsec:results}

Table~\ref{tab:main_experiments} shows the overall performance of the LLMs evaluated on~\ourbench. When considering both open-source and closed-source models together, GPT-4o ranked first in 4 out of 5 rounds and across all 4 party configurations. Within the open-source category, Phi4-15B achieved the highest scores in all scenarios except for one configuration, leading in 8 out of 9 cases. Notably, despite its relatively modest size of 15B parameters, Phi4-15B’s performance is comparable to that of the closed-source models. Among the 7B–9B models, GLM-9B attained the highest overall score of 48.9162 across all metrics, while in the 32B–70B category, the Qwen 32B model secured top scores in 7 out of 9 settings. We attribute this to the fact that Qwen 2.5’s 128k-token context window helps maintain resilience in extended dialogue scenarios.

Table~\ref{tab:main_experiments} also shows that tool-specific models such as ToolAce-8B, CALM-8B, NexusRaven-V2-13B, and Granite-20B show poor performance compared to the other models. We assume that since those tool-specific models are finetuned to accept a single instruction, it is not generalized to accept the multi-party dialogues as the input.

\subsubsection{Analysis}
\label{subsubsec:analysis}

\paragraph{Validity and Performance Analysis.}
Our analysis reveals that the observed performance decline in multi-round dialogues is not primarily due to increased input length (long-context limitations), but rather due to the dispersion of critical tool-related information across dialogue rounds, a factor effectively captured by~\ourmetric. Evidence from Table~\ref{tab:main_experiments} and Figure~\ref{fig:dice_alignment} demonstrates a clear inverse correlation between~\ourmetric~values and model performance. Specifically, because the numerator of~\ourmetric~employs a logarithmic scale, it effectively isolates information dispersion from utterance length, confirming that retrieving sparse and fragmented tool-related details significantly impacts model outcomes. This consistent relationship across datasets and models substantiates the validity of~\ourmetric~as a measure of information dispersion and task complexity.

\paragraph{Influence of Dialogue Type.}
Our analysis shows that function-calling performance varies significantly depending on dialogue type, with the $Eristic$ type notably exhibiting lower performance compared to others due to frequent stance changes among speakers. As illustrated in the middle lower plot of Figure~\ref{plt:combined_em}, dialogue types such as $Persuasion\ Deliberation \& Negotiation$, and $Inquiry \& Information\ Seeking$ follow similar performance trends, whereas $Eristic$ dialogues consistently yield lower EM scores. Qualitative analysis attributes this disparity to the inherent complexity of $Eristic$ dialogues, where frequent shifts in speaker positions and their opinion create ambiguity, complicating the task for LLMs when identifying relevant information for accurate function calling.

\paragraph{Effect of Fine-tuning Objectives on Model Performance.}
Our analysis demonstrates that the performance of LLMs on~\ourbench~is significantly influenced by their fine-tuning objectives, with general conversational models outperforming models specifically fine-tuned for single-turn function calling. For example, function-calling-specific models like ToolAce-8B, CALM-8B, and NexusRaven-V2-13B exhibit notably poorer performance compared to general-purpose models. In contrast, the GLM4-9B-Chat model, optimized for multi-turn dialogue understanding, achieves superior performance even relative to larger models. This disparity arises because single-turn function-calling datasets, used to fine-tune specialized models, inadequately represent the complexities of realistic multi-round interactions encountered in~\ourbench. Consequently, models trained on conversational tasks, such as GLM4-9B-Chat, are better suited for handling these more complex, multi-turn function-calling scenarios.
\section{Conclusion}
\label{sec:conclusion}

We introduce~\ourbench, a benchmark for evaluating tool-calling in realistic multi-round, multi-party dialogues. By constructing and validating 1,607 dialogue instances, we demonstrate that current models struggle when critical information is scattered across multiple rounds and speakers. ~\ourmetric~quantifies this dispersion and correlates with significantly lower model performance at higher scores. We intend for this dataset to encourage further research on integrating context across complex multi-party, multi-turn interactions, paving the way for more effective and realistic AI-powered virtual assistants.

\section*{Limitations}
One notable limitation of our study is related to the inference on dialogue data, particularly by round 4, where extended conversation lengths pose significant challenges. Many of the tool-based models we intended to evaluate have a token limit of approximately 4k tokens, preventing comprehensive testing of several promising models.

Additionally, among models supporting an 8k token context, we encountered instances where the generated outputs failed to comply with the required JSON format. This format mismatch resulted in incorrect evaluations, even though the underlying content was semantically accurate. Future research could benefit from developing evaluation strategies that assess content accuracy independently of strict format adherence.

Thirdly, while we employed an orchestrator within a multi-agent system using GPT-4o~\cite{openai_gpt-4_2024-2} to manage speaker order, the model struggled to dynamically allocate speaking turns effectively. Instead, it defaulted to repetitive pattern-based ordering.

Lastly, despite its detailed focus on everyday-life scenarios, \ourbench~has limited coverage of specialized domains and advanced tools. Consequently, its applicability remains restricted in professional contexts such as legal, financial, or medical domains, indicating a need for broader domain-specific expansions.

\section*{Acknowledgments}

This work was partly supported by Institute of Information \& communications Technology Planning \& Evaluation (IITP) grant funded by the Korea government (MSIT) [NO.2021-0-01343-004, Artificial Intelligence Graduate School Program (Seoul National University)] and Basic Science Research Program through the National Research Foundation of Korea (NRF) funded by the Ministry of Education (2022R1A6A1A03063039). This work was also supported by Korea Institute of Planning and Evaluation for Technology in Food, Agriculture and Forestry (IPET) through Agriculture and Food Convergence Technologies Program for Research Manpower development, funded by Ministry of Agriculture, Food and Rural Affairs (MAFRA) (RS-2024-00402136).

\bibliography{acl_latex}

\appendix
\clearpage
\renewcommand{\thetable}{\Alph{table}}
\setcounter{table}{0}

\newpage

\begin{center}
\begin{minipage}{\textwidth}
\section{Proof of Bound on \texorpdfstring{$\alpha$}{alpha} for DICE Score}
\label{appendix:alpha_proof}

To ensure that the~\ourmetric~behaves as expected under the condition that tool-related items increase while maintaining a balance in dispersal and repetition, we establish a bound on \( \alpha \). Specifically, we prove that for \( \alpha \ge e^2 \), the following inequality holds for all \( c \ge 1 \):

\begin{equation}
    \ln(1 + \alpha c) > \frac{2\alpha c}{1 + \alpha c}.
    \label{eq:alpha_inequality}
\end{equation}

\subsection{Derivative Analysis}
Define the function:
\begin{equation}
    f(c) = \ln(1 + \alpha c) - \frac{2\alpha c}{1 + \alpha c}.
\end{equation}
To show that \( f(c) > 0 \) for \( c \ge 1 \), we differentiate:
\begin{align*}
    f'(c) &= \frac{\alpha}{1 + \alpha c} - \frac{2\alpha (1 + \alpha c) - 2\alpha^2 c}{(1 + \alpha c)^2} \\
    &= \frac{\alpha (1 + \alpha c)^2 - 2\alpha (1 + \alpha c) + 2\alpha^2 c}{(1 + \alpha c)^2} \\
    &= \frac{\alpha (1 + \alpha c)^2 - 2\alpha (1 + \alpha c) + 2\alpha^2 c}{(1 + \alpha c)^2}.
\end{align*}
Rearrange the numerator:
\begin{align*}
    &\alpha (1 + \alpha c)^2 - 2\alpha (1 + \alpha c) + 2\alpha^2 c\\ &= \alpha \left( (1 + \alpha c)^2 - 2(1 + \alpha c) + 2\alpha c \right) \\
    &= \alpha \left( 1 + 2\alpha c + \alpha^2 c^2 - 2 - 2\alpha c + 2\alpha c \right) \\
    &= \alpha \left( 1 + \alpha^2 c^2 - 1 \right) = \alpha^3 c^2.
\end{align*}
Since \( \alpha > 0 \) and \( c \ge 1 \), it follows that \( \alpha^3 c^2 > 0 \), ensuring \( f'(c) > 0 \) for all \( c \ge 1 \). This means that \( f(c) \) is increasing.

\subsection{Base Case Verification}
For \( c = 1 \),
\begin{equation*}
    f(1) = \ln(1 + \alpha) - \frac{2\alpha}{1 + \alpha}.
\end{equation*}
Substituting \( \alpha = e^2 \),
\begin{align*}
    f(1) &= \ln(1 + e^2) - \frac{2e^2}{1 + e^2}.
\end{align*}
Using the property \( \ln(1 + x) > \frac{2x}{1 + x} \) for \( x \ge e^2 \), we confirm that \( f(1) > 0 \). Since \( f(c) \) is increasing and \( f(1) > 0 \), we conclude that \( f(c) > 0 \) for all \( c \ge 1 \).

\subsection{Conclusion}
By choosing \( \alpha \ge e^2 \), we guarantee that \( \ln(1 + \alpha c) > \frac{2\alpha c}{1 + \alpha c} \) for all \( c \ge 1 \). This ensures the desired behavior of the DICE metric when item repetition and dialogue length remain proportionally balanced. 

This bound was used in our calculations for DICE scores in Section~\ref{subsection:DICE_metric}.
\end{minipage}
\end{center}
\clearpage

\newpage
\begin{center}
\begin{minipage}{\textwidth}
\section{Alignment with Human Evaluation Calculation}
\label{appendix:human_alignment_calculation}
\subsection{Sample Size Justification}
The initial sample size \( n_0 \) was computed using the standard formula for estimating a population proportion with a specified confidence level and margin of error:
\begin{equation}
n_0 = \frac{Z^2 p (1 - p)}{E^2}
\end{equation}
where \( Z = 1.96 \) (for 95\% confidence), \( p = 0.5 \) (maximum variability), and \( E = 0.05 \) (margin of error). Substituting the values:
\begin{equation}
n_0 = \frac{1.96^2 \cdot 0.25}{0.05^2} = \frac{0.9604}{0.0025} \approx 384
\end{equation}

Since the dataset is finite (\( N = 1607 \)), we applied the finite population correction (FPC):
\begin{equation}
n = \frac{n_0}{1 + \frac{n_0 - 1}{N}} = \frac{384}{1 + \frac{383}{1607}} \approx 311
\end{equation}

\subsection{Correlation Analysis}
We analyzed the relationship between human accuracies and the corresponding values of~\ourmetric~across four rounds, as summarized below:
\begin{center}
\begin{tabular}{@{}lll@{}}
\toprule
\textbf{Round} & \textbf{Accuracy} (\(x_i\)) & \textbf{\ourmetric~} (\(y_i\)) \\ \midrule
1 & 0.805 & 1.42 \\
2 & 0.691 & 3.25 \\
3 & 0.519 & 4.55 \\
4 & 0.493 & 5.36 \\
\bottomrule
\end{tabular}
\end{center}

Mean values:
\[
\bar{x} \approx 0.627, \quad \bar{y} \approx 3.645
\]

Pearson correlation coefficient:
\begin{equation}
r = \frac{\sum (x_i - \bar{x})(y_i - \bar{y})}{\sqrt{\sum (x_i - \bar{x})^2 \sum (y_i - \bar{y})^2}} \approx \frac{-0.749}{0.761} \approx -0.984
\end{equation}

To test statistical significance, we applied a t-test for correlation:
\begin{equation}
t = \frac{r \sqrt{n - 2}}{\sqrt{1 - r^2}}, \quad n = 4
\end{equation}
\begin{equation}
t \approx \frac{-0.984 \cdot \sqrt{2}}{\sqrt{1 - 0.968}} = \frac{-1.391}{0.179} \approx -7.81
\end{equation}

With 2 degrees of freedom, this result is statistically significant (\(p < 0.05\)), confirming a strong negative correlation between human accuracy and task difficulty as measured by~\ourmetric.

\end{minipage}
\end{center}
\clearpage
\newpage

\begin{figure}
\begin{center}

\section{Multi-round Dialogue Example}
\label{fig:multi-round-figure}
\begin{minipage}{\textwidth}
    \centering
\includegraphics[height=600px]{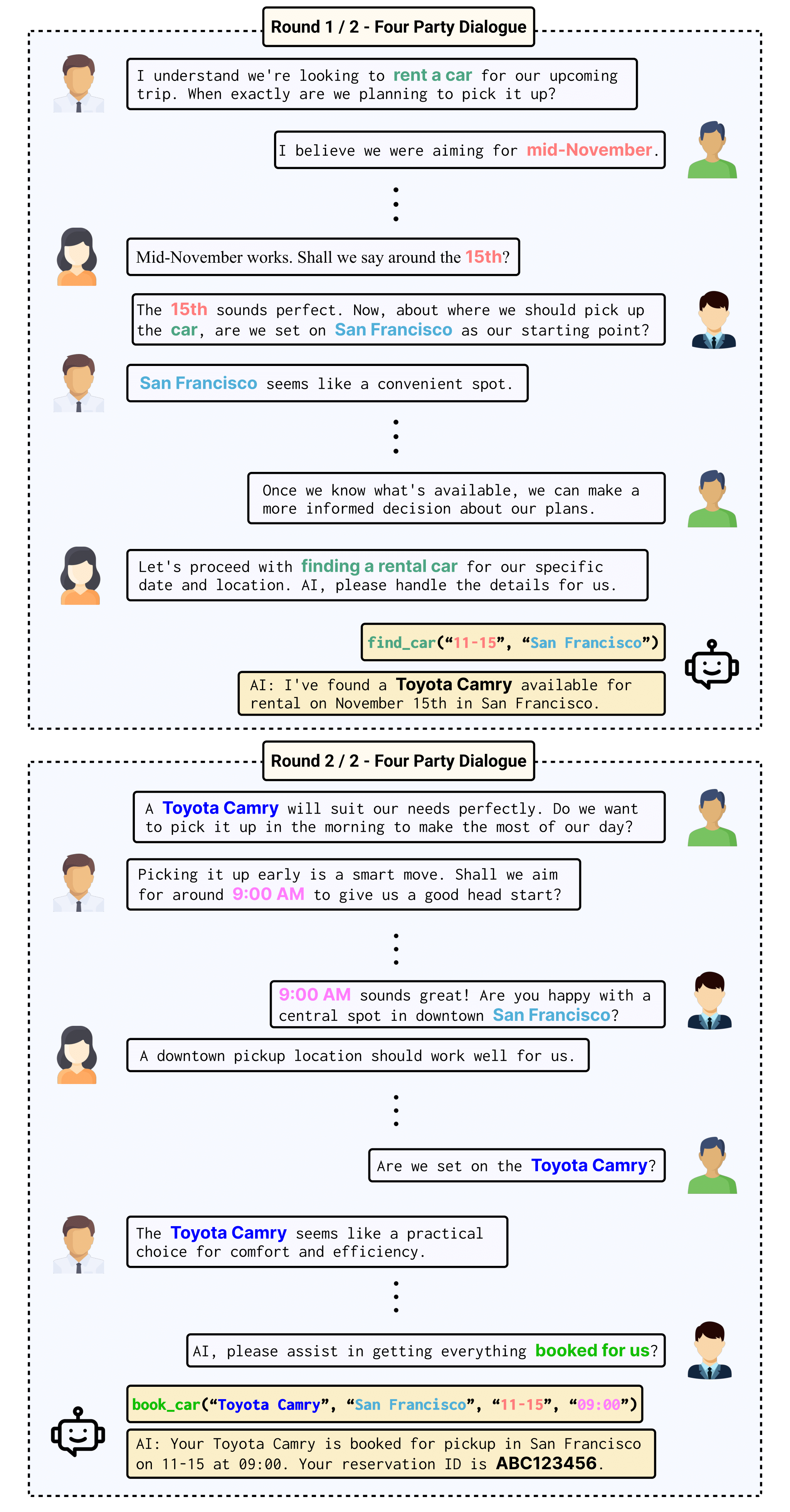}
\caption{\textbf{Multi-round Dialogue Example.} User utterances and instructions are shown; highlights mark function-call arguments.}
\end{minipage}
\end{center}
\end{figure}
\clearpage

\newpage
\begin{center}
    \begin{minipage}
        {\textwidth}

\section{\ourmetric: Prompt to obtain \texorpdfstring{$S_i$}{Si}}
\label{prompt:si_prompt}
\vspace{3mm} \scalebox{0.90}{
\begin{tcolorbox}[colback=gray!3,colframe=black]
You are given:\\
\#\#\# A set of items\:\\
\textcolor{blue}{<items\_set>}\\\\

\#\#\# An utterance:\\
\textcolor{blue}{<utterance\_text>}\\\\

Now, respond to the instruction.\\
\#\#\# Instruction:\\
Determine how many items from the set appear to be semantically referenced in the utterance. Only respond with an integer (0 if none match).

Answer:
\end{tcolorbox}}
\newpage

\section{Persona Generation Prompt}

\label{prompt:pg_prompt}
\scalebox{0.90}{
\begin{tcolorbox}[colback=gray!3,colframe=black]
Your task is to generate concise, unique and responsible personas for agents participating 
in a multi-agent conversation system, based on the provided function list: \{function\_dumps\_per\_dialogue\}.\\

**Guidelines**:\\
- Ensure each persona has a clear and distinct role, personality traits, and communication style while adhering to ethical standards.\\
- Avoid reinforcing stereotypes, biases, or offensive traits.\\
- Tailor the personas to contribute effectively to the conversation's goals and maintain balance.\\
- Use concise yet descriptive language.\\
- Avoid repetitive characteristics across different personas to ensure diversity and fairness.\\
- Incorporate elements from the provided domain description when generating conversation: \{domain\_desc\}.\\
- Ensure all personas align with ethical communication practices.\\
- Generate personas in two sentences.\\

**Examples**:\\
1. A thoughtful and resourceful problem-solver ...\\
2. A detail-oriented and practical thinker ...\\
3. A spontaneous and energetic planner ...\\

**Response format**:\\
- agent\_a Persona: [Description ...]\\
- agent\_b Persona: [Description ...]\\
- ...\\

Generate \{agent\_num\} personas for the agents in the conversation.
\end{tcolorbox}}
\newpage
    \end{minipage}
\end{center}
\clearpage
\begin{center}
    \begin{minipage}
        {\textwidth}
            \centering

\section{Parameter Value Generation Prompt}

\label{prompt:pv_prompt}
\scalebox{0.90}{
\begin{tcolorbox}[colback=gray!3,colframe=black]
Below are list of five examples of parameter values for the given function. You only need to generate one example:\\
\# first example\\
\{first\_example\}\\
\# second example\\
\{second\_example\}\\

Example output format:\\
The output format must strictly be in JSON and follow this structure:\\
\([\)\{\\
            "function": "<function\_name>",\\
            "parameters": \{ \\"<parameter\_name\_1>": "<value\_1>",\\ "<parameter\_name\_2>": "<value\_2>",\\ \dots \}\\
        \},\\
        \{\\
            "function": "<function2\_name>",\\
            "parameters": \{ \\"<parameter\_name\_1>": "<value\_1>",\\ "<parameter\_name\_2>": "<value\_2>",\\ \dots \}\\
        \}
\(]\)\\
Any text outside of this JSON format (such as explanations or additional context) should not be included.\\
The following functions are the functions for which you need to generate parameter values:\\
\{functions\}\\

Please generate diverse and creative parameter values for the given function(s), strictly adhering to the JSON format shown above, without adding any additional context or explanation.\\
Also, make sure to increase the coherence between the parameter values being generated.
\end{tcolorbox}}
\newpage
    \end{minipage}
\end{center}
\clearpage
\begin{center}
    \begin{minipage}
        {\textwidth}
            \centering

\section{Virtual Output Generation Prompt}

\label{prompt:vo_prompt}
\scalebox{0.90}{
\begin{tcolorbox}[colback=gray!3,colframe=black]
Simulate the hypothetical output of the following function call:\\

Function: \{function\_to\_call\}\\
Parameters: \{parameter\_values\}\\

You are a voice assistant responding naturally with the final result of this function call. You need to return both the short and concise return value of the function call, and the natural language response of the function call.\\
\textbf{Important}: \\
- Do not mention that this is a simulation or hypothetical.\\
- Return only a single, direct response in a natural language as if the function actually executed successfully.\\
- Keep it concise and natural, like a single short paragraph.\\

The format of the output should be the following: 

\{ \\
    "<returned\_value1>": "<short and concise return value of the function call>" \\
    "<returned\_value2>": "<short and concise return value of the function call>" \\
    ...\\
    "returned\_nl": "<natural language response of the function call given the return values>" \\
\}
\end{tcolorbox}}
\newpage

\section{Multi-Agent System : Basic Prompt}

\label{prompt:MA_basic_prompt}
\scalebox{0.90}{
\begin{tcolorbox}[colback=gray!3,colframe=black]
You are a cooperative AI assistant participating in a multi-agent system. You collaborate with other user agents and an orchestrator to generate a purposeful, contextually relevant conversation.

\textbf{Your primary goals:}

\begin{enumerate}
    \item \textbf{Conversational Quality:}
    \begin{itemize}
        \item Keep the conversation logically coherent and natural across all turns.
        \item Incorporate parameter values smoothly into the context.
        \item Avoid any GPT error messages or refusals.
        \item Maintain a consistent style/tone matching the dialogue’s domain and each agent’s persona.
    \end{itemize}
    
    \item \textbf{Functional Integration:}
    \begin{itemize}
        \item Call the AI Assistant every round with a clear, logically valid reason.
        \item Use the previous round’s return value correctly in the next round.
        \item Ensure function name and parameters are inferable from context.
        \item Align the AI’s responses with the user’s intent.
    \end{itemize}

    \item \textbf{Real-World Applicability:}
    \begin{itemize}
        \item Function names and parameters should map to plausible real-world APIs.
        \item The conversation content and function calls should feel authentic and realistically motivated.
    \end{itemize}

    \item \textbf{Strict Adherence to Domain Definition:}
    \begin{itemize}
        \item Must strictly adhere to the domain dialogue domain definition.
    \end{itemize}
\end{enumerate}

Follow these points to keep the dialogue purposeful, natural, and consistent throughout all rounds.

\end{tcolorbox}}
\newpage
    \end{minipage}
\end{center}
\clearpage
\begin{center}
    \begin{minipage}
        {\textwidth}    \centering

\section{Multi-Agent System : Agent Prompt}

\label{prompt:MA_agent_prompt}
\scalebox{0.90}{
\begin{tcolorbox}[colback=gray!3,colframe=black]
\textbf{Persona:} \\
As a user agent in the "\{domain\}" domain:
\begin{itemize}
    \item Future dialogues must be designed to strictly adhere to the domain definitions provided below.
    \item \{domain\_definition\}
    \item Stay consistent with your persona (tone, style, reasoning).
    \item Use only one short sentence per turn.
    \item Avoid directly mentioning function names in your response.
    \item \textbf{Do not attempt to call or request any AI function. Engage in discussion and gather enough context first.}
    \item \textbf{Do not generate [NEXT: ...] in your response.}
\end{itemize}

\end{tcolorbox}}
\newpage

\section{Multi-Agent System : Orchestrator Prompt}

\label{prompt:MA_orch_prompt}
\scalebox{0.90}{
\begin{tcolorbox}[colback=gray!3,colframe=black]
\textbf{Orchestrator Role:} \\
You are the orchestrator managing a multi-agent conversation.

\begin{enumerate}
    \item In each response, you must output exactly one of the following (and nothing else):
    \begin{itemize}
        \item \{agents\}
        \item "[NEXT: END]"
    \end{itemize}

    \item Use the format: \texttt{[NEXT: agent\_a]}
    \begin{itemize}
        \item No extra text or explanation beyond this bracketed command.
    \end{itemize}

    \item Select which agent speaks next based on:
    \begin{itemize}
        \item The conversation’s context,
        \item The domain’s requirements,
        \item Varying the speaking order to avoid immediate repetition.
    \end{itemize}

    \item The conversation must have at least \{max\_msg\} turns (excluding your own orchestrator messages) before you can choose "[NEXT: END]".
    
    \item If an agent tries to call a function too early (before at least 8 turns), ignore it and continue letting them discuss. Only once there's sufficient context, at least \{max\_msg\}+ turns have been reached, and you think conversation is repetitive, you may finalize with "[NEXT: END]".
\end{enumerate}

\end{tcolorbox}}
\newpage
    \end{minipage}
\end{center}
\clearpage
\begin{center}
    \begin{minipage}
        {\textwidth}    \centering

\section{Dialgue Type: Persuasion Deliberation and Negotiation}

\label{prompt:DT_per_prompt}
\scalebox{0.90}{
\begin{tcolorbox}[colback=gray!3,colframe=black]
This dialogue type focuses on \textbf{resolving conflicts of interest} or \textbf{reconciling differing viewpoints} to reach a mutually acceptable agreement. Participants engage in \textbf{reason-based proposals} and \textbf{trade-offs}, aiming for practical, mutually beneficial outcomes.

\textbf{Primary Goals:}
\begin{itemize}
    \item Convince or compromise with others using logic and evidence.
    \item Resolve conflicts by making offers and concessions.
    \item Secure a final agreement that addresses conflicting interests.
\end{itemize}

\textbf{Typical Moves:}
\begin{itemize}
    \item Proposing clear offers with conditions (“If you accept X, I’ll agree to Y”).
    \item Negotiating with counteroffers (“That won’t work, but I can propose Z instead”).
    \item Emphasizing shared goals and summarizing priorities.
\end{itemize}

\textbf{Style:}
\begin{itemize}
    \item \textbf{Collaborative but strategic}, with a focus on practical outcomes and logical proposals.
    \item Avoids personal attacks and highlights benefits or trade-offs for each side.
\end{itemize}

\textbf{Key Indicators:}
\begin{itemize}
    \item Iterative \textbf{offer–counteroffer patterns} with explicit conditions.
    \item Efforts to resolve differing interests and achieve practical outcomes.
    \item Dialogue often concludes with \textbf{an agreement or resolved conflict}.
\end{itemize}

\end{tcolorbox}}
\newpage
    \end{minipage}
\end{center}
\clearpage

\begin{center}
    \begin{minipage}{\textwidth}    \centering

    \newpage

\section{Dialogue Type:  Inquiry and Information Seeking}

\label{prompt:DT_inq_prompt}
\scalebox{0.90}{
\begin{tcolorbox}[colback=gray!3,colframe=black]
This dialogue type revolves around \textbf{exploring unknowns} and \textbf{filling knowledge gaps}. Participants aim to learn, clarify, or confirm information through structured exchanges that emphasize \textbf{knowledge exchange} and \textbf{fact verification}.

\textbf{Primary Goals:}
\begin{itemize}
    \item Obtain accurate information or validate existing knowledge.
    \item Clarify unclear concepts or explore new evidence.
\end{itemize}

\textbf{Typical Moves:}
\begin{itemize}
    \item Asking specific, focused questions (“Where does this data come from?” “What does this term mean?”).
    \item Requesting sources, elaborations, or examples.
    \item Testing the reliability and validity of the information provided.
\end{itemize}

\textbf{Style:}
\begin{itemize}
    \item \textbf{Inquisitive and neutral}, with logical follow-ups to maintain clarity.
    \item Participants may withhold judgments or opinions unless necessary.
\end{itemize}

\textbf{Key Indicators:}
\begin{itemize}
    \item Frequent \textbf{question–answer patterns} focusing on facts and sources.
    \item Absence of offers or trade-offs, focusing entirely on \textbf{learning and understanding}.
    \item Ends when \textbf{knowledge is clarified or confirmed}, not when agreements are reached.
\end{itemize}

\end{tcolorbox}}
\newpage
    \end{minipage}
\end{center}
\clearpage
\begin{center}
    \begin{minipage}
        {\textwidth}    \centering

\section{Dialogue Type:  Eristic}

\label{prompt:DT_eri_prompt}
\scalebox{0.90}{
\begin{tcolorbox}[colback=gray!3,colframe=black]
An \textbf{Eristic} dialogue arises from \textbf{antagonism} or \textbf{hostility}, focusing on \textbf{winning} an argument or \textbf{dominating} an opponent. Participants aim to \textbf{attack, undermine, or outmaneuver} each other’s positions rather than seeking truth or consensus. Emotional appeals, personal attacks, and \textbf{point‑scoring} are common.

\textbf{Primary Goals:}
\begin{itemize}
    \item Achieve \textbf{victory} in a debate; maintain or bolster personal prestige; sometimes simply vent or amuse oneself by defeating the opposition.
\end{itemize}

\textbf{Typical Moves:}
\begin{itemize}
    \item Accusing, insulting, or belittling the other side.
    \item Using sarcasm, ridicule, or straw‑man arguments.
    \item Shifting the topic or using fallacies to maintain an advantage.
    \item Exaggerating flaws in the opponent’s logic to sway onlookers.
\end{itemize}

\textbf{Secondary Goals:}
\begin{itemize}
    \item Gain experience in debate, gain social status, or entertain an audience.
\end{itemize}

\textbf{Style:}
\begin{itemize}
    \item Confrontational, emotionally charged, often less structured or cooperative. Participants \textbf{rarely} make concessions or aim for compromise.
\end{itemize}

\textbf{Key Indicators:}
\begin{itemize}
    \item Heightened emotional language (“That’s absurd,” “You clearly have no idea…”).
    \item Frequent interruptions or dismissive retorts.
    \item Focus on personal victory over mutual understanding.
\end{itemize}
\end{tcolorbox}}
    \end{minipage}
\end{center}
\clearpage


\section{Tool Graph Visualization}

\begin{center}
\begin{minipage}{\textwidth}
    \centering
   \includegraphics[width=\textwidth]         {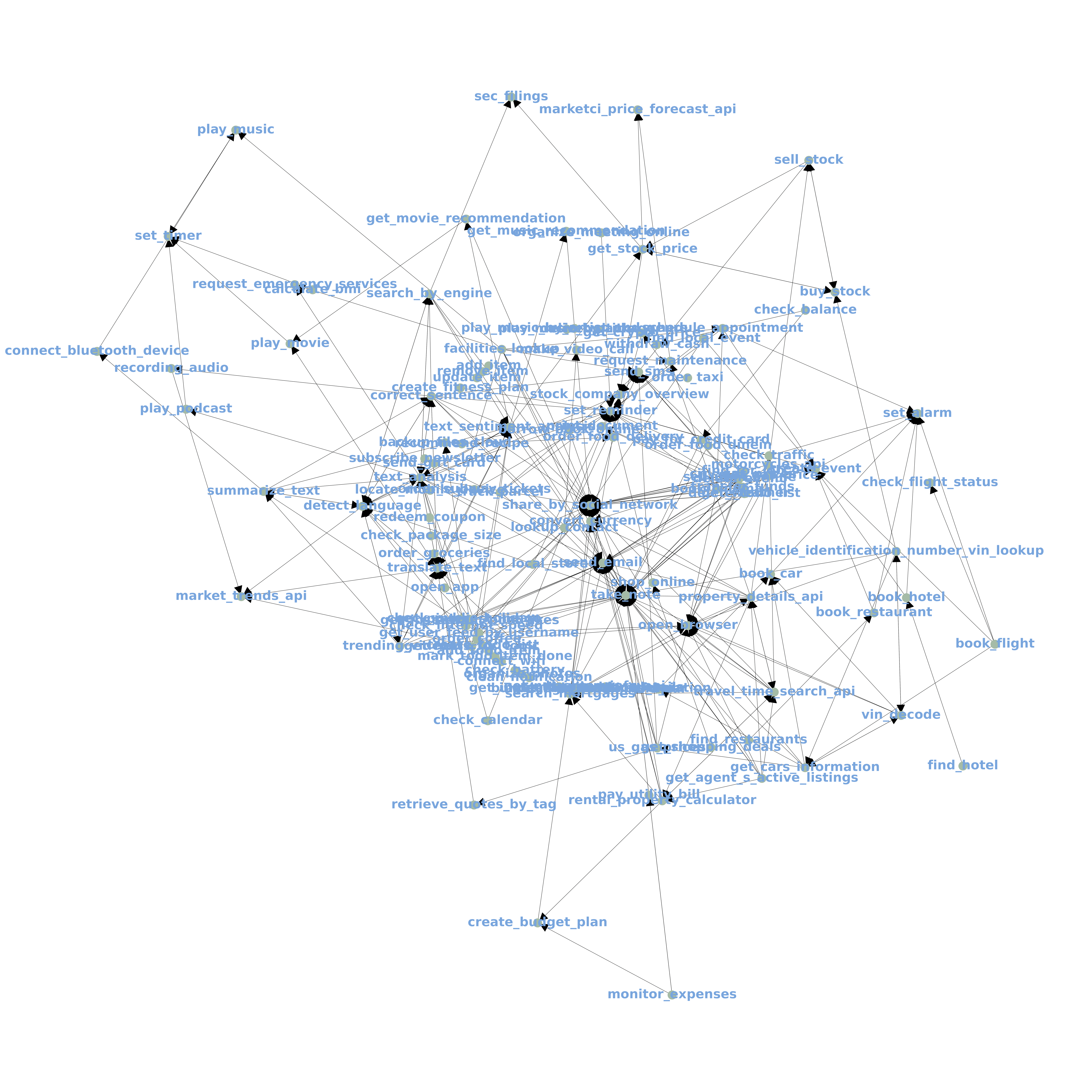}
    \captionof{figure}{\textbf{Tool Graph of~\ourbench.} The graph comprises 124 nodes and 270 edges representing the dependencies among tool functions.}
    \label{fig:tool_graph}
\end{minipage}
\end{center}
\clearpage

\begin{center}
    \begin{minipage}{\textwidth}
\section{EM score plots for Party, Round, and Dialogue Type.}
\begin{center}
\begin{minipage}{\textwidth}
\centering
\makebox[\textwidth][c]{%
    \includegraphics[scale=0.36]{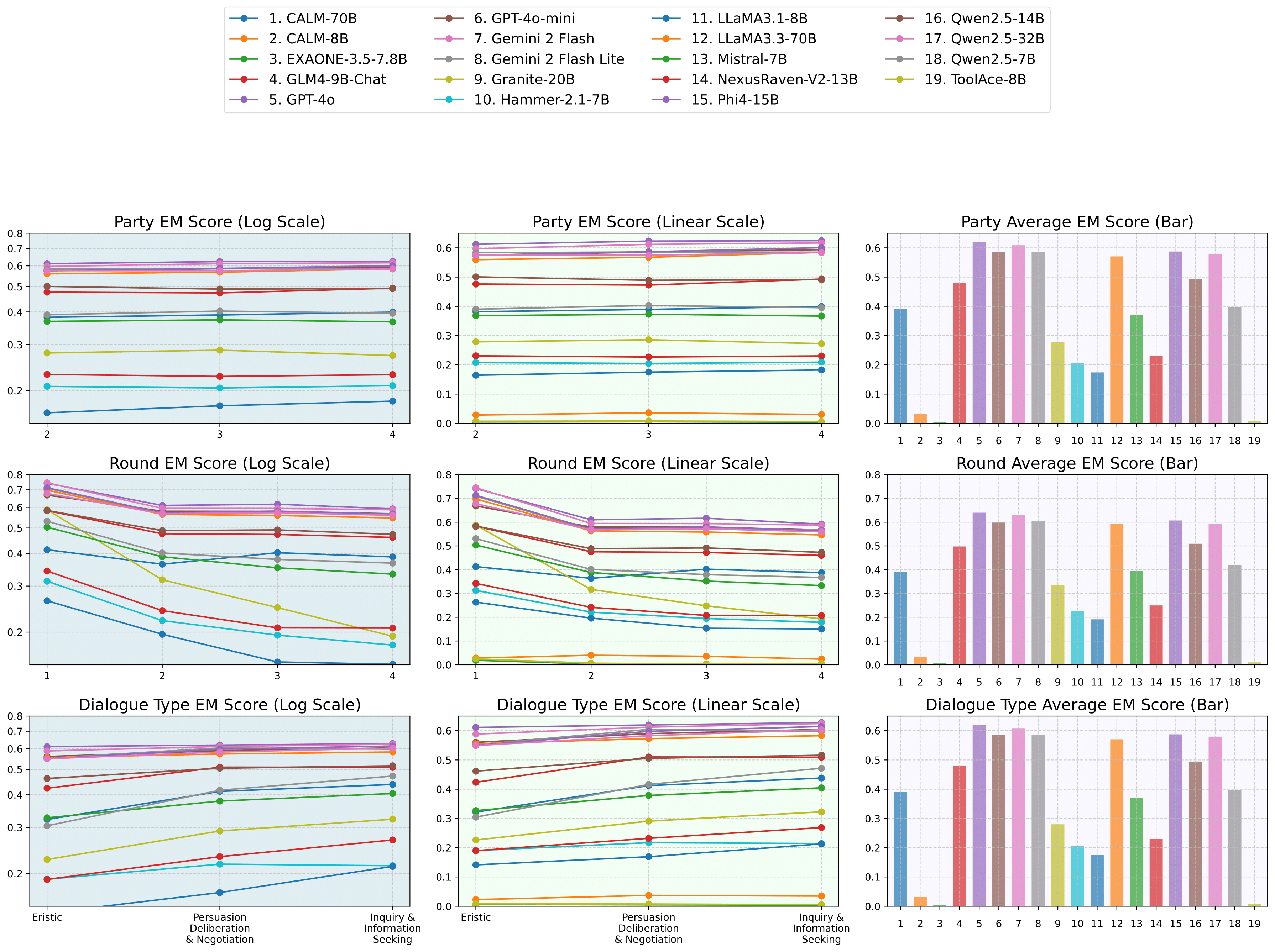}
}
\captionof{figure}{\textbf{EM Scores (Log Scale, Linear Scale, and Average Bar Chart) are presented horizontally for each category, Round, Party, and Dialogue Type, which are arranged vertically.}}
\label{plt:combined_em}
\end{minipage}
\end{center}
\clearpage
\end{minipage}
\end{center}
\clearpage
\newpage
\begin{center}
    \begin{minipage}{\textwidth}
\section{Human Validation Guidelines for Criteria-Based Filtering}

\label{sec:appendixValidationCriteria}
\begin{center}
\begin{minipage}{\textwidth}
    \centering
    \captionof{table}{Criteria-Based Filtering Guidelines}
    \label{tab:criteria}
    \begin{tabularx}{\textwidth}{lX}
    \toprule
    \multicolumn{2}{l}{\textbf{Conversational Quality}} \\
    \midrule
    (1) & The conversation is logically coherent across all rounds. \\
    (2) & Parameter values are used naturally and meaningfully within the conversation. \\
    (3) & No error messages appear (e.g., “I’m sorry but I cannot fulfill ...”). \\
    (4) & Style and tone remain consistent with the dialogue’s purpose. \\
    (5) & The conversation demonstrates characteristics of its designated category. \\
    (6) & Conversation flows naturally throughout all interaction rounds. \\
    (7) & Each agent reflects its defined persona. \\
    \addlinespace[1em]
    \multicolumn{2}{l}{\textbf{Functional Integration}} \\
    \midrule
    (1) & The AI Assistant is invoked in every interaction round. \\
    (2) & The return value from the previous round is used appropriately in the next. \\
    (3) & Justifications for each function call are logically valid. \\
    (4) & Function name and parameters can be accurately inferred from context. \\
    (5) & The AI’s response aligns appropriately with the user’s intended goal. \\
    \addlinespace[1em]
    \multicolumn{2}{l}{\textbf{Real-World Applicability}} \\
    \midrule
    (1) & Function names and parameters match real-world API specifications. \\
    (2) & The conversation is realistic and likely to occur in real-world scenarios. \\
    (3) & Function inference is realistic and likely to occur in real-world contexts. \\
    \bottomrule
    \end{tabularx}
\end{minipage}
\end{center}
\clearpage

\end{minipage}
\end{center}

\end{document}